\documentclass{article}

     \PassOptionsToPackage{round}{natbib}

     \usepackage[final]{neurips_2024}

\usepackage[utf8]{inputenc} %
\usepackage[T1]{fontenc}    %
\usepackage[hyperfootnotes=false]{hyperref}       %
\usepackage{url}            %
\usepackage{booktabs}       %
\usepackage{amsfonts}       %
\usepackage{nicefrac}       %
\usepackage{microtype}      %
\usepackage{xcolor}         %

\usepackage{graphicx} 
\usepackage[ruled,noend,linesnumbered]{algorithm2e}
\usepackage[acronym]{glossaries}

\SetKwComment{Comment}{/* }{ */}

\usepackage{amsmath,amsfonts,bm}

\def\eqref#1{equation~\ref{#1}}

\def\1{\bm{1}}

\def\rvtheta{{\boldsymbol{\theta}}}
\def\rvphi{{\boldsymbol{\phi}}}

\def\rva{{\mathbf{a}}}

\def\rve{{\mathbf{e}}}

\def\rvg{{\mathbf{g}}}

\def\rvs{{\mathbf{s}}}

\def\rvsigma{{\boldsymbol{\sigma}}}
\def\rvmu{{\boldsymbol{\mu}}}

\def\rmSigma{{\mathbf{\Sigma}}}
\def\rmTau{{\boldsymbol{\mathcal{T}}}}

\def\vs{{\bm{s}}}

\DeclareMathAlphabet{\mathsfit}{\encodingdefault}{\sfdefault}{m}{sl}
\SetMathAlphabet{\mathsfit}{bold}{\encodingdefault}{\sfdefault}{bx}{n}

\newcommand{\E}{\mathbb{E}}

\newacronym{cbim}{CB-IM}{competence-based intrinsic motivation}
\newacronym{cpm}{CPM}{Competence Progress Motivation}
\newacronym{diayn}{DIAYN}{Diversity is All You Need}
\newacronym{rl}{RL}{reinforcement learning}
\newacronym{gcrl}{GCRL}{goal-conditioned reinforcement learning}
\newacronym{lp}{LP}{Learning Progress}
\newacronym{vic}{VIC}{Variational Intrinsic Control}
\newacronym{knn}{KNN}{$k$-nearest neighbours}
\newacronym{sd}{$SD$}{Spread Density}
\newacronym{sgd}{SGD}{stochastic gradient descent}
\newacronym{knnf1}{KNN-$F_1$}{$k$-nearest neighbours $F_1$}
\newacronym{bmi}{BMI}{behavioural mutual information}
\newacronym{im}{IM}{intrinsic motivation}
\newacronym{mlp}{MLP}{multi-layer perceptron}
\newacronym{sac}{SAC}{soft actor-critic}
\newacronym{gmm}{GMM}{Gaussian mixture model}
\newacronym{dp}{DP}{Diversity Progress}
\newacronym{curious}{CURIOUS}{Continual Universal Reinforcement learning with Intrinsically mOtivated sUbstitutionS}
\newacronym{tsne}{t-SNE}{$t$-distributed stochastic neighbor embedding}
\newacronym{credit}{CRediT}{Contributor Role Taxonomy}

\glsunset{vic}
\glsunset{diayn}
\glsunset{curious}
\glsunset{cpm}

\title{Diversity Progress for Goal Selection in Discriminability-Motivated RL}

\author{%
    Erik M. Lintunen, Nadia M. Ady\textsuperscript{*}, Christian Guckelsberger\textsuperscript{*} \\
    Department of Computer Science, Aalto University, Finland \\
    \texttt{\{erik.lintunen,nadia.ady,christian.guckelsberger\}@aalto.fi} \\
    \textsuperscript{*}Joint senior authors, order determined alphabetically
}

\begin{document}

\maketitle

\begin{abstract}
  Non-uniform goal selection has the potential to improve the reinforcement learning (RL) of skills over uniform-random selection.
  In this paper, we introduce a method for learning a goal-selection policy in intrinsically-motivated goal-conditioned RL: ``Diversity Progress'' (DP).
  The learner forms a curriculum based on observed improvement in discriminability over its set of goals.
  Our proposed method is applicable to the class of discriminability-motivated agents, where the intrinsic reward is computed as a function of the agent's certainty of following the true goal being pursued. This reward can motivate the agent to learn a set of diverse skills without extrinsic rewards.
  We demonstrate empirically that a DP-motivated agent can learn a set of distinguishable skills faster than previous approaches, and do so without suffering from a collapse of the goal distribution---a known issue with some prior approaches.
  We end with plans to take this proof-of-concept forward.
\end{abstract}

\section{Introduction}

Intrinsically-motivated learning has been studied extensively in the \gls{rl} literature (see recent reviews by \citealp{colas2022autotelic,aubret2023information,lidayan2024bamdp}).
Here, we focus on intrinsically-motivated skill acquisition, often referred to as 
\glspl{cbim}, an area of control problems requiring multiple skills. How to learn a diverse set of skills is a key subproblem of \glspl{cbim} \citep[p.~1161]{colas2022autotelic}. In answer, one class of \glspl{cbim} uses rewards computed as functions of the agent’s certainty of following the goal it chose to pursue, that is, the goal's discriminability. It is postulated in existing work that the more discriminable a set of goals is, the more diverse we expect the skills to be in terms of observed behaviour (see Section~\ref{sec:bmi}). Many such discriminator-based models select goals uniformly during training; yet, learning a distribution over goals has the potential to speed up learning (see Section~\ref{sec:related}).

Our contribution is threefold. (1) We present \gls{dp} (Section~\ref{sec:diversityprogress}), a method for learning a goal-selection policy prioritising goals based on the observed learning progress over a set of goals. (2) We complement the formalism with empirical findings (Section~\ref{sec:experiments}), suggesting that a \gls{dp}-motivated agent can improve existing discriminator-based \glspl{cbim} by speeding up the learning of a diverse set of skills. (3) We detail plans for taking this proof-of-concept forward (Section~\ref{sec:futurework}).

\section{Background}

\subsection{Discriminability-motivated RL}
\label{sec:bmi}

Our method falls within the framework of \gls{gcrl}, in which a \textit{skill} is a policy paired with a goal, which can be identified with a goal-conditioned reward function (see Appendix~\ref{appendix:gcrl}). We focus on a class of intrinsic rewards used to maximise \gls{bmi} (termed by \citealp[p.~2]{hansen2020fast}; e.g., \citealp{gregor2016variational,warde-farley2019unsupervised,eysenbach2019diversity,baumli2021relative,laskin2022unsupervised}), though our method could be used with any discriminator-based \gls{cbim}. These rewards approximate the mutual information between the goal-defining variable, $\rvg$, and some function of the trajectory drawn from the corresponding skill, $f(\rmTau_{\pi_{\rvg}})$. Typically, $f$ maps a trajectory to a single state, but some formulations map to multiple states (e.g., \citealp{gregor2016variational}, use the initial and final states). Following \citet[p.~4]{hansen2020fast}, the \gls{bmi} objective can be defined as the maximisation of:
\begin{align}
\label{eq:bmi}
I(\rvg; f(\rmTau_{\pi_{\rvg}})) := H(\rvg) - H(\rvg \mid f(\rmTau_{\pi_{\rvg}})),
\end{align}
where $H$ represents Shannon entropy. However, mutual information can be notoriously hard to compute \citep[p.~1]{hjelm2019learning}, and a popular approach is to maximise a variational lower bound:
\begin{align}\label{eq:bmiapprox}
\tilde{I}(\rvg; f(\rmTau_{\pi_{\rvg}})) \geq H(\rvg) - \mathbb{E}_{\rvg\sim p(\rvg),\rmTau_{\pi_{\rvg}}\sim \pi(\rvg)}\left[\log q(\rvg \mid f(\rmTau_{\pi_{\rvg}}))\right],
\end{align}
where $q$ is an arbitrary variational distribution \citep[p.~2]{barber2003im}. Given $f(\rmTau_{\pi_{\rvg}})$, $q$ defines a probability distribution over goals. That is, a probability model, $q$, predicts, for each skill, the probability that the skill has induced the observations---thus the model is canonically known as a \textit{discriminator}---and is typically used to compute the reward. Successfully discriminating skills in the observation space requires the agent to observe distinct regions of the state space, encouraging the agent to learn a set of diverse behaviours. If a goal is not discriminable based on observations, two or more skills are producing overlapping behaviours (and therefore, the skills lack diversity); conversely, if a goal is discriminable, then the corresponding skill is inducing trajectories unique to that skill.

\subsection{Learning Progress}
\label{sec:learningprogress}

\Gls{lp} \citep[e.g.,][]{oudeyer2007intrinsic, schmidhuber2010formal} is designed ``to select goals that are of learning-optimal difficulty with respect to [the agent's] current capabilities'' \citep[p.~39]{lintunen2024advancing}. 
\citet[pp.~270--271]{oudeyer2007intrinsic}
formulate \gls{lp} as an intrinsic reward measuring how much the agent has improved in some prediction over a window of time. For any decision that affects learning, the \gls{lp} at time $t+1$ can be computed for each option, indexed by $n$:
\begin{align}\label{eq:lp_n}
\mathrm{LP}_n(t+1) := \overline{e}_n(t+1-\tau) - \overline{e}_n(t+1),
\end{align}
given an offset, $\tau$, of the comparison, in time steps.
A smoothing hyperparameter $\eta$ adds robustness to stochastic fluctuations, so the average error $\overline{e}_n$ is defined as the mean over the smoothing window:
\begin{align}\label{eq:lperrors}
\overline{e}_n(t+1-\tau) := \frac{1}{\eta+1} \sum_{i=0}^\eta e_n(t+1-\tau-i),
\end{align}
where $e_n(t+1)$ is the difference between some prediction and the true observation made due to choosing option $n$. Originally, predictions were of state observations \citep[p.~271]{oudeyer2007intrinsic}. \Gls{lp} has since been extended to measures of competence beyond prediction error (\citealp{colas2019curious}).

\section{Related work}
\label{sec:related}

\paragraph{\gls{bmi}-maximising \glspl{cbim}} In \gls{vic} \citep{gregor2016variational}, one of the earliest \gls{bmi}-related approaches, the agent learns a categorical distribution over goals, $\rvg \sim p(\rvg)$---like our method. \Gls{vic} reinforces $p(\rvg)$ to maximise the same \gls{bmi}-based intrinsic reward as is used for learning the skills (p.~4). However, \citet[p.~6]{eysenbach2019diversity} showed that in \gls{vic} the probability mass collapses to a handful of skills. In response, \citeauthor{eysenbach2019diversity} fixed $p(\rvg)$ as uniform in \gls{diayn}, so all skills receive, in expectation, equal training signal. A side effect is inefficient use of training samples: the curriculum is not optimised for improving discriminability. \Gls{dp} aims to train more efficiently, via a learned goal-selection policy that could be used with any \gls{bmi}-maximising \glspl{cbim}. 
\citet[Table~2, p.~10]{laskin2022unsupervised} list several discriminator-based approaches, noting key differences in reward formulations.

\paragraph{\gls{lp}-based \glspl{cbim}} \Gls{lp} has been used to motivate goal selection in a variety of ways \citep[p.~1185]{colas2022autotelic}. Learning $p(\rvg)$ can significantly improve the speed of learning a set of skills \citep[pp.~260--261]{stout2010competence}. Our method formulates goal selection as a bandit problem rewarded using \gls{lp}, much like the approach used by \gls{curious} \citep{colas2019curious}, one such \gls{lp}-based \gls{cbim}. This formulation is in contrast with \gls{vic}, which sets an \gls{rl} problem, conditioning its goal-selection policy on a state. However, while \gls{dp} selects between single goals in the full sensory space, \gls{curious} supports the provision of predefined subspaces, which the authors call modules. The \gls{curious} agent learns a distribution over the modules, learning to prioritise subspaces of goals it is increasingly or decreasingly successful in reaching (success is determined by some binary function).

\section{Diversity Progress}
\label{sec:diversityprogress}

Our method motivates the agent to prioritise goals that provide most progress in the learning of the discriminator, over all skills. That is, if pursuing one goal improves the discriminability of multiple skills, it is reflected positively in the goal-selection probabilities. We call this \acrlong{dp}.

At a given time, $t+1$, the agent's prediction error for a given goal, $\rvg \sim p(\rvg)$, is defined by:
\begin{align}\label{eq:dpprederrors}
e_\rvg(t+1) :=
    \begin{cases}
        1-q(\rvg \mid \rvs_{t+1}), & \text{if } \rvg \text{ corresponds to the skill being followed} \\
        q(\rvg \mid \rvs_{t+1}), & \text{otherwise,}
    \end{cases}
\end{align}
where $e_\rvg(t+1)$ is an element of $\rve(t+1) \in \mathbb{R}^{|\mathcal{G}|}$, a vector composed of the error, at $t+1$, for each goal, and $q$ is a discriminator. Then, the errors over an epoch of fixed time length $T$ form a $T \times |\mathcal{G}|$ matrix. Given hyperparameters \textit{offset}, $\tau$, and \textit{smoothing}, $\eta$, (see Section~\ref{sec:learningprogress}), we use \autoref{eq:lperrors} to compute the most recent average errors, $\overline{\rve}(t+1)$, and the average errors from $\tau$ time steps earlier, $\overline{\rve}(t+1-\tau)$. Following \autoref{eq:lp_n}, we compute the \gls{lp} for each goal, with goal $\rvg$ corresponding to option $n$. Then, the \gls{lp} values are averaged over goals to compute the \acrlong{dp}:
\begin{align}\label{eq:dpaverageovergoals}
\overline{\textit{DP}}(t+1) := \frac{1}{|\mathcal{G}|} \sum_{\rvg \in \mathcal{G}} \overline{\textnormal{e}}_\rvg(t+1-\tau) - \overline{\textnormal{e}}_\rvg(t+1).
\end{align}
In this way, the mean progress \textit{over all goals} is attributed to the goal being pursued, since that goal determines the actions taken. For each goal, $\rvg$, $\overline{\textit{DP}}$ is updated after executing the corresponding skill. The goal-selection probabilities, $p(\rvg)$, are computed via softmax. For the algorithm see Appendix~\ref{appendix:algorithms}.

\section{Experiments}
\label{sec:experiments}

For details of implementation and evaluation see Appendices \ref{appendix:implementation} and \ref{appendix:evaluation}, respectively.

\subsection*{RQ1. Does the probability mass of the goal distribution collapse?}
\label{sec:rq1collapse}

\begin{figure}[h]
\begin{center}
\includegraphics[width=1\linewidth]{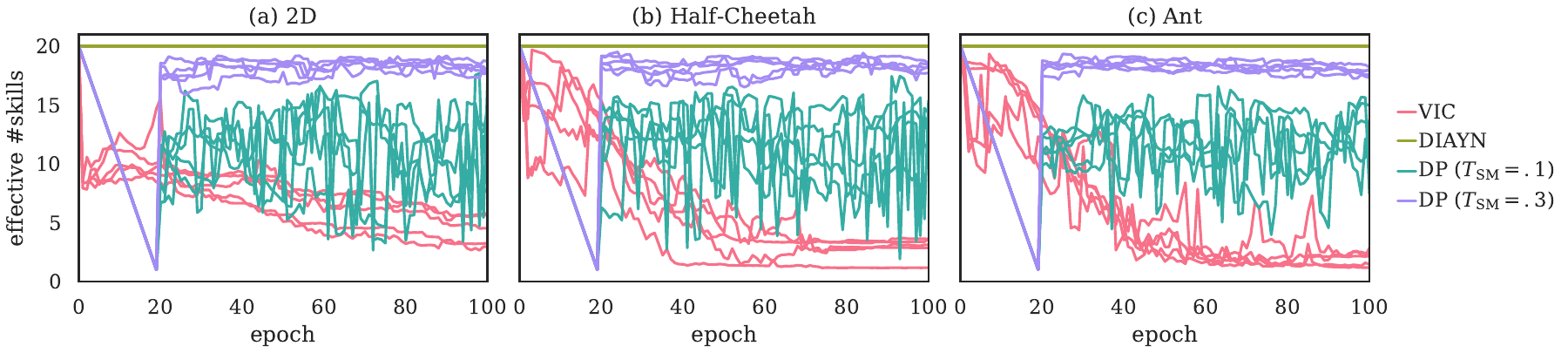}
\end{center}
\vspace{-0.4cm}
\caption{\textbf{The effective number of skills} over training time in three environments. The agent is learning $20$ skills. We compare \gls{vic}, \gls{diayn}, and \gls{dp} with two different softmax temperatures (0.1 and 0.3) determining how greedy the policy is. The linear decline of the effective number of skills for epochs up to the number of skills is due to \gls{dp}'s initialisation, that is, randomly selecting goals without replacement (see Algorithm~\ref{alg:diversity}, ll.~6--10). Results from five random seeds; each line is a seed.}
\label{fig:effectivenskills}
\end{figure}

To learn a diverse set of skills, it is material to train a substantial number of skills. \citet[p.~6]{eysenbach2019diversity} showed that, with \gls{vic}'s goal-selection method, the effective number of skills (see Appendix~\ref{appendix:effectivenskills}) collapses to a handful. The method by \citet{eysenbach2019diversity}, \gls{diayn}, therefore selects goals uniformly. Does \gls{dp} avoid a collapse like \gls{vic}'s? Figure~\ref{fig:effectivenskills} shows the effective number of skills over training time for all three methods in three environments (see Appendix~\ref{appendix:environments}). With the fixed $p(\rvg)$ in \gls{diayn}, its effective number of skills is constant---equal to $|\mathcal{G}|$. \Gls{dp} does not collapse, and, additionally, we have control over the effective number of skills via the softmax temperature.

\subsection*{RQ2. What are the effects of \gls{dp} on the dynamics of goal selection?}
\label{sec:rq2dpeffects}

\begin{figure}[h]
\begin{center}
\includegraphics[width=1\linewidth, trim=0.27cm 0 0 0, clip]{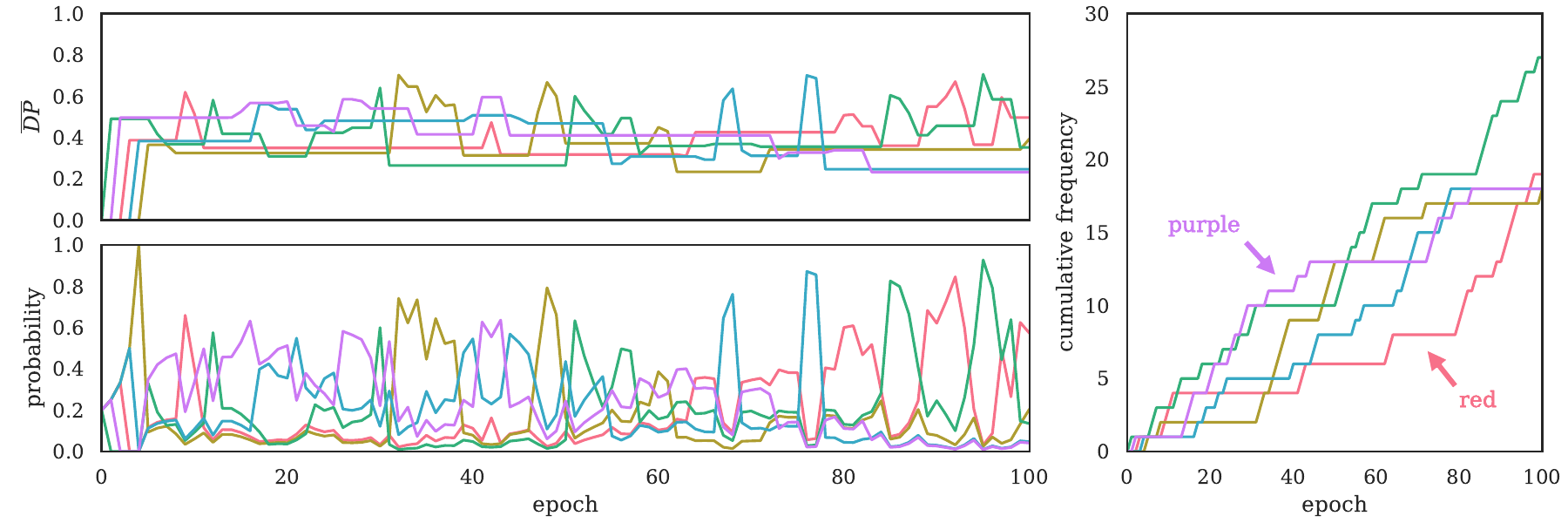}
\end{center}
\vspace{-0.4cm}
\caption{\textbf{The effects of \gls{dp} on goal selection} over training time in the Half-Cheetah environment. The agent is learning five skills, each shown in a different colour. \textit{Upper left:} \gls{dp} values, $\overline{\textit{DP}}$; updated for the current skill at the end of an epoch. \textit{Lower left:} goal-selection probabilities after the softmax transformation. \textit{Right:} cumulative frequencies of goal selection. The initial trend for epochs up to the number of skills is due to \gls{dp}'s initialisation, where goals are selected randomly without replacement (see Algorithm~\ref{alg:diversity}, ll.~6--10). The plotted traces represent data from running a single random seed.}
\label{fig:dpvalues}
\end{figure}

Figure~\ref{fig:dpvalues} shows \gls{dp} values, goal-selection probabilities, and cumulative frequencies (counts of how many times each goal has been selected thus far) for a small number of skills (five). Initially, without evidence on all goals, each goal is selected once. Then, the agent should focus on the ``easiest-to-discriminate'' skills, up until repeated attempts generate less progress than other goals. Goals that are too ``hard'' generate little progress, so should be avoided until other goals are sufficiently mastered.

Consider the purple and red goals. Red initially provided low \gls{dp}, so was largely ignored until it was sufficiently easy (the policy is not completely greedy, so red is still updated occasionally despite its difficulty) or others provide less \gls{dp} in comparison (e.g., purple declining). From the cumulative frequencies of selection, we see the red goal eventually caught up---explained by its increase in \gls{dp}.

\subsection*{RQ3. Does \gls{dp}-guided selection of goals speed up learning distinguishable skills?}
\label{sec:rq3dpadvantages}

\Gls{lp} should prioritise goals that are providing the most improvement to the discriminator. Therefore, we hypothesise that a \gls{dp}-motivated agent learns a set of diverse skills faster than a learner sampling goals uniformly. We probe this by visualising trajectories in a dimension-reduced feature space to find out whether trajectories from early training are distinguishable based on their mean observations (\autoref{fig:tsne100epochs}; details in Appendix~\ref{appendix:tsne}). Compared to random skills (no learning) and skills learned with uniform goal selection (\gls{diayn}), the set of skills learned with \gls{dp} is distinguishable earlier.

\begin{figure}[h]
\begin{center}
\includegraphics[width=1\linewidth]{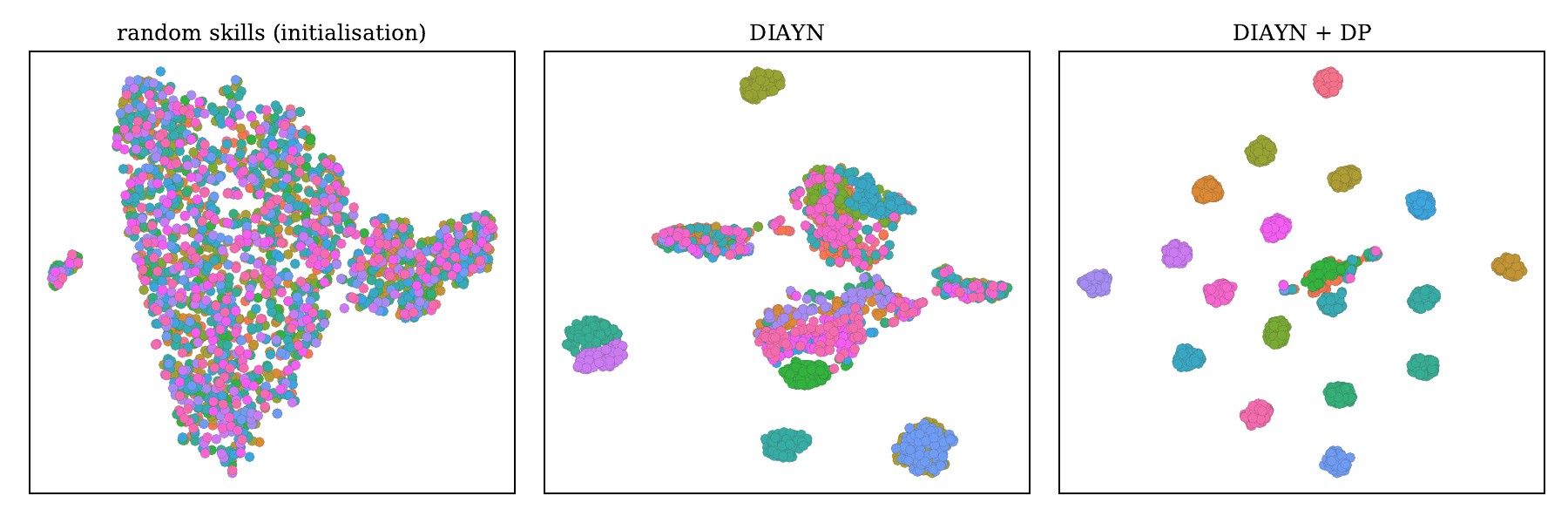}
\end{center}
\vspace{-0.4cm}
\caption{\textbf{A dimension-reduced feature space (\acrshort{tsne}) for 20 skills} in the Half-Cheetah environment. For each skill (one colour), the $100$ data points represent i.i.d. draws of trajectories.
\textit{Left:} randomly initialised skills with no training. \textit{Middle, Right:} trajectories sampled after $100$ epochs of training. Note both \gls{dp} and \gls{diayn} eventually learn distinguishable skills. Experiment details in  Appendix~\ref{appendix:tsne}.}
\label{fig:tsne100epochs}
\end{figure}

\section{Conclusion and future work}
\label{sec:futurework}

We propose \gls{dp} as a method for learning a goal-selection policy in discriminability-motivated \gls{rl}, prioritising goals based on overall improvements in discriminability. We have shown in three experiments, that: (1) a \gls{dp}-motivated agent learns a distribution over goals without the probability mass collapsing; (2) the \gls{dp} values motivate goal selection with respect to observed \gls{lp}; and (3) with \gls{dp}, an agent can learn a diverse set of skills in less training time than with uniform-random selection.

In future work, we aim to better understand how different factors affect goal selection. We plan to test other intrinsic rewards combining discriminability with \gls{lp}, including absolute \gls{lp}, where the agent also attends to goals that it is forgetting (i.e., skills decreasing discriminability). Different entropy regularisation regimes may benefit diversity in terms of increased state-space coverage but make discrimination of skills harder. Following \citet{eysenbach2019diversity}, the utility of \gls{dp} can be tested on transfer learning and hierarchical tasks (pp.~6,~7--8), comparing with other discriminator-based \glspl{cbim}, and on a range of environments including non-episodic and stochastic ones. However, evaluating open-ended learning is notoriously difficult. When the aim is to learn a diverse set of ``interesting'' skills, what is interesting is typically subjectively defined by the researchers and the community, and not clearly designed to maximise a well-defined objective \cite[p.~1168]{colas2022autotelic}. We plan to improve the evaluation of skill diversity by considering different diversity metrics, and use our findings to quantitatively evaluate \gls{dp} against the current state-of-the-art \gls{bmi}-maxisiming \glspl{cbim}.

\section*{Author contributions}
\label{sec:credit}

We follow the \acrfull{credit} introduced by \citet{brand2015beyond}.
Conceptualisation: initial equals EML, NMA; evolution equals EML, NMA, CG.
Data curation: EML.
Formal analysis: EML.
Funding acquisition: CG.
Investigation: EML.
Design of methodology: lead EML; supporting NMA, CG.
Creation of models: leads EML, NMA; supporting CG.
Software: EML.
Resources: CG.
Oversight: equals NMA, CG.
Leadership responsibility: CG.
Visualisation: lead EML; supporting NMA, CG.
Writing: original draft EML; review \& editing equals EML, NMA, CG.

\section*{Acknowledgements and disclosure of funding}
\label{sec:acks}

EML and CG received financial support from the Research Council of Finland (NEXT-IM, grant 349036) and NMA from the Helsinki Institute for Information Technology.
We acknowledge the computational resources provided by the Aalto Science-IT project.
We thank the Aalto Scientific Computing team, particularly Simo Tuomisto, Mira Salmensaari, and Hossein Firooz, for their support.
For their time, ideas, and feedback on the topic of measuring skill diversity, we thank Perttu Hämäläinen, Marlos Machado, Sebastian Berns, Nam Hee Kim, and Luigi Acerbi.
We thank the reviewers for their service and for providing useful future-looking feedback.

\bibliographystyle{plainnat} %
\bibliography{refs} %

\newpage
\appendix

\section{Goal-conditioned RL}
\label{appendix:gcrl}

\Acrlong{gcrl} (for a review see \citealp{liu2022goal}) extends the standard definition of a reward function \citep[p.~1]{kaelbling1993learning} to be conditioned on goals:
\begin{align}
R_G: \mathcal{S} \times \mathcal{A} \times \mathcal{S} \times \mathcal{G} \rightarrow \mathbb{R},
\end{align}
where $\mathcal{S}$ is \text{the set of possible states}, $\mathcal{A}$ is \text{the set of actions available to the agent}, and $\rvg \in \mathcal{G}$, known as a \textit{goal}, or a \textit{goal-defining variable}, is a parameter to the reward function (cf., \citealp[p.~1165]{colas2022autotelic}; \citealp[p.~6]{aubret2023information}). Note that $R_G(\rvs, \rva, \rvs', \rvg)$ is a random variable, even for fixed $\rvs, \rvs'\in \mathcal{S}$, $\rva \in \mathcal{A}$, and $\rvg \in \mathcal{G}$. Then, a \textit{skill} is a policy paired with a goal, optimising for the return according to the reward conditioned on that goal. For example, the goal-defining variable, $\rvg$, can be an index \citep[e.g.,][]{eysenbach2019diversity} or an element drawn from a learned distribution \citep[e.g.,][]{nair2018visual}. Its role is simply to indicate which reward function the agent is aiming to maximise.

Goals can be viewed as ``a set of \emph{constraints} ... that the agent seeks to respect'' \citep[p.~1165, emphasis in original]{colas2022autotelic}. While the most immediate intuition of a goal is often as a desired state for the agent to reach (e.g., \citealp[p.~1]{kaelbling1993learning}; \citealp[p.~2]{schaul2015universal}), the formalism allows for a more general set of constraints on behaviour \citep[pp.~22--23]{lintunen2024advancing}. In effect, any behaviour that can be defined by attempting to maximise some reward function on the environment can be formulated as a goal--skill pairing.

\section{Algorithm}
\label{appendix:algorithms}

\begin{algorithm}[H]\label{alg:diversity}
\DontPrintSemicolon
\caption{\Acrlong{dp} for goal selection}
\SetKwFunction{UpdateDP}{UpdateDP}
Set hyperparameters: smoothing $\eta$, offset $\tau$, softmax temperature $T_\mathrm{SM}$, number of goals $|\mathcal{G}|$. \\
Initialise skills $\pi_\rvtheta$, with parameters $\rvtheta$, and discriminator $q_\rvphi$, with parameters $\rvphi$. \\
Initialise uniform, categorical distribution over goals: $p(\rvg) = \mathcal{U}\{1,|\mathcal{G}|\}$. \\
Initialise zero vector for \acrlong{dp}: $\mathbf{dp} = \mathbf{0}$, where $\mathbf{dp} \in \mathbb{R}^{|\mathcal{G}|}$. \\
Initialise environment with initial state $\rvs_0$ according to the environment. \\
\For(\Comment*[f]{Select each goal once to set $p(\rvg)$}){$epoch \leftarrow 1$ \KwTo $|\mathcal{G}|$}{ 
  Sample goal $\rvg \sim p(\rvg)$ without replacement. \\
  \UpdateDP{$\rvg$}}
Normalise $\mathbf{dp}$ to a probability mass function via softmax: $\exp(\mathbf{dp}/T_\mathrm{SM})/\sum_j \exp(\textnormal{dp}_j/T_\mathrm{SM})$. \label{line:softmax} \\
Update goal distribution: $p(\rvg) \leftarrow \mathbf{dp}$. \label{line:updatep}\\
\While{training}{
  Sample goal $\rvg \sim p(\rvg)$.\\ %
  \UpdateDP{$\rvg$} \\
  Normalise $\mathbf{dp}$ and update $p(\rvg)$ as in lines \ref{line:softmax}--\ref{line:updatep}.
}
\vspace{1.5\baselineskip}
  \SetKwProg{Pn}{function}{:}{\KwRet}
  \Pn{\UpdateDP{$\rvg$}}{
          \For{$t \leftarrow 1$ \KwTo steps\_per\_epoch}{
    Sample action $\rva_t \sim \pi_\rvtheta(\rva_t \mid \rvs_t,\rvg)$ from skill associated with $\rvg$. \label{line:sampleaction}\\
    Step environment: $\rvs_{t+1} \sim p(\rvs_{t+1} \mid \rvs_t,\rva_t)$. \\
    Compute prediction errors: $1-q_\rvphi(\rvg \mid \rvs_{t+1})$ if $\rvg$ is goal being pursued else $q_\rvphi(\rvg \mid \rvs_{t+1})$. \\
    Update discriminator ($\rvphi$) with optimiser of choice. \label{line:updateq}\\
    Use reward / algorithm / optimiser of choice to learn goal-conditioned policy. \label{line:learnpolicy}
  }
  Compute most recent mean errors: $\overline{\rve}(t+1) = \frac{1}{\eta+1} \sum_{i=0}^\eta \rve(t+1-i)$. \label{line:recenterrors}\\
  Compute mean errors from $\tau$ time steps earlier: $\overline{\rve}(t+1-\tau) = \frac{1}{\eta+1} \sum_{i=0}^\eta \rve(t+1-\tau-i)$. \\
  Compute \acrlong{dp}: $\overline{\textit{DP}}(t+1) = \frac{1}{|\mathcal{G}|} \sum_{\rvg \in \mathcal{G}} \overline{\rve}(t+1-\tau) - \overline{\rve}(t+1)$. \\
  Update \acrlong{dp} for current goal $\rvg$: $\textnormal{dp}_\rvg \leftarrow \overline{\textit{DP}}(t+1)$. \label{line:updatedp}
  }
\end{algorithm}

In our implementation (see Appendix~\ref{appendix:implementation}), we follow \gls{diayn} \citep{eysenbach2019diversity} on lines \ref{line:updateq}--\ref{line:learnpolicy}.

\section{Implementation}
\label{appendix:implementation}

\subsection{Learning objective}

Our implementation is based on \gls{diayn}, so following \citet[p.~3]{eysenbach2019diversity} we wish to ensure: (1) that states---not actions---are used to distinguish skills; and (2) that the agent is maximising the entropy of its policies. This leads to the following instantiation of the \gls{bmi} objective (see Section~\ref{sec:bmi} for \gls{bmi}; see \citealp[Equations~1--2, p.~4,]{eysenbach2019diversity} for the full derivation):
\begin{align}
\label{eq:bmivariant}
\mathcal{F}(\rvtheta) &:= I(\mathcal{S}; \mathcal{G}) + H(\mathcal{A} \mid \mathcal{S}) - I(\mathcal{A}; \mathcal{G} \mid \mathcal{S}) = H(\mathcal{A} \mid \mathcal{S}, \mathcal{G}) - H(\mathcal{G} \mid \mathcal{S}) + H(\mathcal{G}).
\end{align}

The objective function is a variational lower bound on \autoref{eq:bmivariant} (see \citealp[p.~4]{eysenbach2019diversity}):
\begin{align}
\label{eq:bmivariantapprox}
\mathcal{F}(\rvtheta, \rvphi) := \underbrace{H(\mathcal{A} \mid \mathcal{S},\mathcal{G})}_\text{(a)} + \underbrace{\E_{\rvg\sim p(\rvg), \rvs\sim \pi_\rvtheta(\rvg)}\left[\log q_\rvphi(\rvg \mid \rvs)-\log p(\rvg)\right]}_\text{(b)},
\end{align}

where $q_\rvphi$ is a discriminator, specifically a \gls{mlp}, parametrised by $\rvphi$.

\subsection{Optimiser and reward function}

In optimising its policy to maximise \autoref{eq:bmivariantapprox}, the agent relies on \gls{sac} \citep{haarnoja2018soft}, an off-policy maximum entropy actor-critic algorithm. In maximum entropy \gls{rl}, where learning a stochastic policy is desirable, the standard \gls{rl} objective of maximising expected return is augmented with an entropy maximisation term, so the optimiser takes care of the entropy term (a) \citep[cf.,][p.~4]{eysenbach2019diversity}. The expectation (b) is maximised with the intrinsic reward function
\begin{align}
\label{eq:reward}
R^i_G := \log q_\rvphi(\rvg \mid \rvs) - \log p(\rvg).
\end{align}
This expresses that the agent is rewarded for its ability to discriminate the skill being followed (see Section~\ref{sec:bmi}). The distribution over goals, $p(\rvg)$, is: (1) categorical; and (2) learned using \gls{dp} (we describe the method for learning a goal-selection policy in Section~\ref{sec:diversityprogress}). At the beginning of an epoch, the agent samples a skill, $\pi_{\rvg \sim p(\rvg)}$, and follows it until termination. That is when $p(\rvg)$ is updated.

\subsection{Action-selection policy}

At the action-selection level, the agent learns a goal-conditioned policy, $\pi_\rvtheta(\rva \mid \rvs,\rvg)$. Following \gls{diayn} (\href{https://github.com/ben-eysenbach/sac/blob/2116fc394749ca745f093a36635a9b253da8170d/examples/mujoco_all_diayn.py#L210-L216}{mujoco\_all\_diayn.py}, ll.~210--216), the policy is a \gls{gmm} with $K$ components, where an \gls{mlp}, parametrised by $\rvtheta$, maps from state--goal pairs to the (log) weight $w_k$, mean vector $\rvmu_k$, and vector $\rvsigma_k$ of (log) standard deviations, i.e., of the diagonal entries of the covariance matrix $\rmSigma_k$, of each Gaussian component:
\begin{align}
\label{eq:gmmpolicy}
\pi_\rvtheta(\rva \mid \rvs, \rvg) := \sum_{k=1}^K w_k \mathcal{N}(\rva; \rvmu_k, \mathrm{diag}(\rvsigma_k)).
\end{align}
The weights $w_1,\cdots,w_K$ are transformed into probabilities for selecting the component (let the parameters of the choice be indexed by $*$) from which to sample an action: $\rva \sim \mathcal{N}_*(\rva; \rvmu_*, \mathrm{diag}(\rvsigma_*))$.

\subsection{Hyperparameters}

We tested various candidates for the \gls{dp} hyperparameters: smoothing, $\eta$, offset, $\tau$, and softmax temperature $T_\mathrm{SM}$. However, since we currently lack a good diversity metric to evaluate differences in performance, the hyperparameter choices in our experiments were mainly based on qualitative observations. Smoothing ranged 100--250 time steps, offset 250--900, and softmax temperature 0.1--0.75. The hyperparameter values used in producing the figures are listed in Table~\ref{table:hyperparameters}. The most noticeable difference was given by varying $T_\mathrm{SM}$. To achieve stability in the level of greediness of the goal-selection policy, we normalised the mean \gls{lp} before computing the mean progress over goals. 

For reference, in the 2D environment, episodes last for 100 time steps, so a goal is sampled every 10 episodes. In the Half-Cheetah environment, episodes last for 1000 time steps, so a new goal is sampled for every episode. In the Ant environment, episodes last for a maximum of 1000 time steps (there is a potential termination condition before the maximum episode length), so a sampled goal may be used starting from partway through an episode and into another episode. For environment details see Appendix~\ref{appendix:environments}.

\begin{table}[h!]
\centering
\caption{The hyperparameters used to generate the figures.}
\begin{tabular}{lllllll}
\toprule
                               &                    & Fig.~\ref{fig:effectivenskills}(a)  & Fig.~\ref{fig:effectivenskills}(b)  & Fig.~\ref{fig:effectivenskills}(c)  & Fig.~\ref{fig:dpvalues} & Fig.~\ref{fig:tsne100epochs} \\
\midrule
smoothing                      & $\eta$           & 250     & 250     & 250     & 250   & 100   \\
offset                         & $\tau$             & 250 & 250 & 750 & 750   & 900   \\
temperature                    & $T_\mathrm{SM}$    & 0.1/0.3 & 0.1/0.3 & 0.1/0.3 & 0.1   & 0.75  \\
number of goals                & $|\mathcal{G}|$ & 20      & 20      & 20      & 5     & 20    \\
number of Gaussian components  & $K$                & 4       & 4       & 4       & 4     & 4     \\
entropy regularisation scaling & $\alpha$           & 0.1     & 0.1     & 0.1     & 0.1   & 0.1   \\
\gls{mlp} number of hidden units         &                    & 32      & 300     & 300     & 300   & 300 \\
\bottomrule
\end{tabular}
\label{table:hyperparameters}
\end{table}
For clarification, the same $K$ and $\alpha$ are used for \gls{diayn} and \gls{diayn}+\gls{dp} across all our experiments. In \autoref{fig:tsne100epochs}, the \gls{dp}-specific hyperparameters ($\eta$, $\tau$, and $T_\mathrm{SM}$) are not needed for \gls{diayn} without \gls{dp}.

\section{Evaluation}
\label{appendix:evaluation}

\subsection{Environments}
\label{appendix:environments}

We carried out experimental comparisons of \gls{dp} against \gls{diayn}, our baseline method, in environments where \gls{diayn} has been shown to perform well, following the advice of \citet[p.~26]{patterson2023empirical}. This choice eases comparison with related work. Specifically, we used three environments tested by \citet{eysenbach2019diversity}: a slightly modified version of their \textit{2D Navigation} environment and two MuJoCo environments \citep{todorov2012mujoco}, namely \textit{Half-Cheetah} and \textit{Ant}. 
The 2D Navigation environment affords diagnostics into the algorithms' function, and with the MuJoCo environments, we represent tasks of increasing complexity in terms of both observation space and degrees of freedom.

\begin{figure}[h]
\begin{center}
\includegraphics[width=0.89\linewidth]{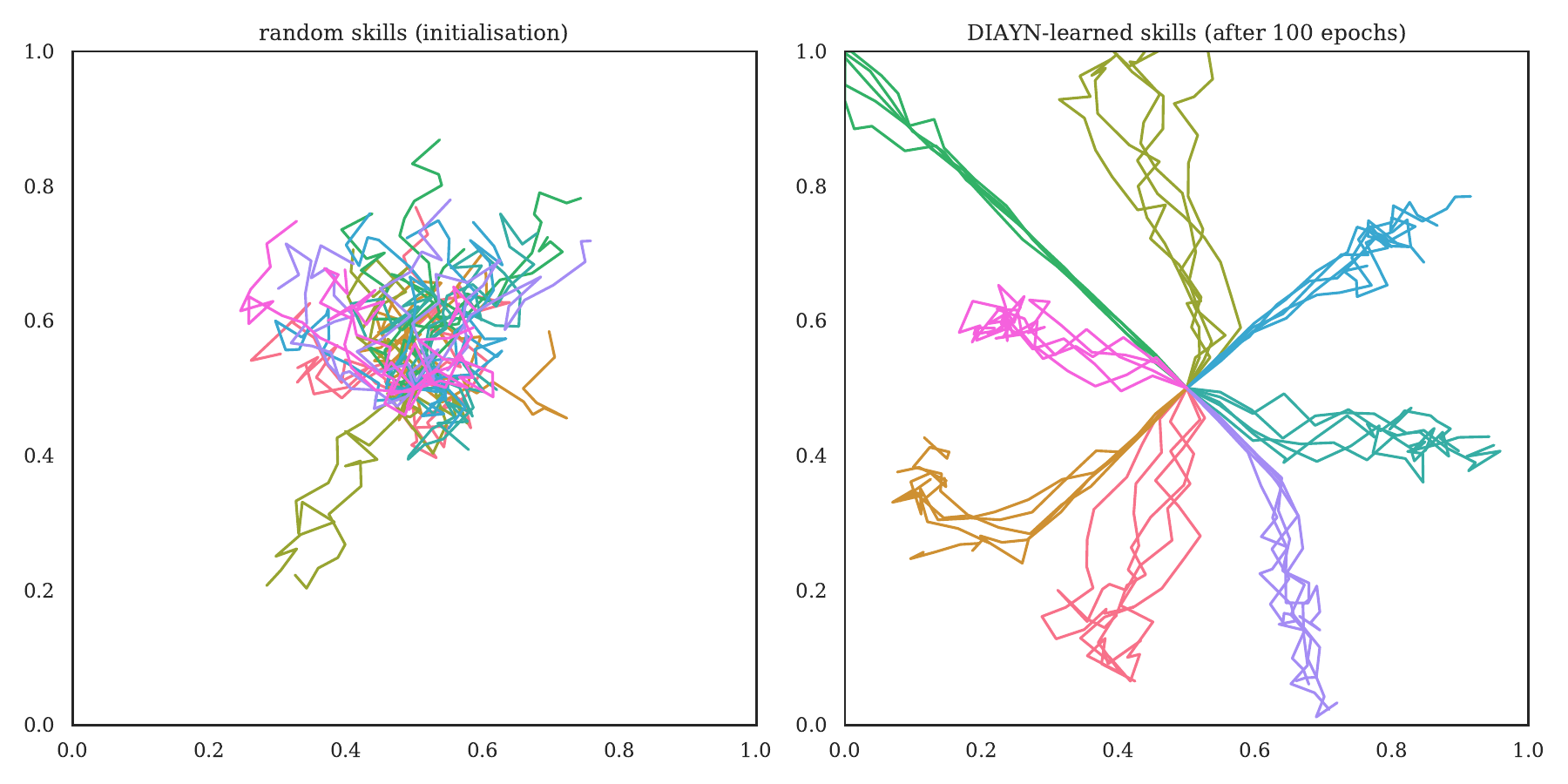}
\end{center}
\vspace{-0.4cm}
\caption{Trajectories drawn from eight stochastic skills in \textbf{our modified version of the 2D Navigation environment} constructed by \citet{eysenbach2019diversity}. \textit{Left:} random skills with no training. \textit{Right:} DIAYN-learned skills. Note, these visualisations are provided for intuition, showing trajectories only 15 steps long. In Figure~\ref{fig:effectivenskills}(a), we included 20 skills and trajectories were 100 steps long. }
\label{fig:2dexample}
\end{figure}

\paragraph{2D Navigation}
Our 2D environment (\autoref{fig:2dexample}) is a slightly modified version of the one constructed by \citet{eysenbach2019diversity}. An agent starts in the center of the unit box $\vs_0 = (0.5, 0.5)$ and observes states $\rvs \in [0,1]^2$.
We modified the action space to $\rva \in [-0.05,0.05]^2$ (whereas the original used larger actions, in $[-0.1,0.1]^2$). The environment is bounded, in that if an action takes the learner outside of the box, they are projected back to the closest point in the support of the observation space.

\paragraph{MuJoCo} HalfCheetah-v1 and Ant-v1 with no modifications \citep{todorov2012mujoco}.

\subsection{Effective number of skills}
\label{appendix:effectivenskills}

Following \citet[Appendix~E.2, p.~18]{eysenbach2019diversity}, we compute the effective number of skills by exponentiating the entropy of the goal-selection policy, quantifying the number of skills that the policy is effectively sampling from at the time of selection. For intuition, consider an example of an occasion in which one goal is providing significantly more \gls{dp} than all other goals. This will decrease the effective number of skills, as the learned distribution over goals is less uniform (favouring that one goal). If the \gls{dp} is similar for all goals, then the distribution is close to uniform and the number of effective skills is close to the total number of skills being learned. With \gls{dp}, we have some control over the effective number of skills via the softmax temperature: higher temperatures lead to less emphasis on the differences in \gls{dp} and thus a more uniform sampling distribution; whereas, with low temperatures, we observe higher variance in the effective number of skills over time. High variance in the effective number of skills is likely desirable, since we wish to favour goals only if they provide more \gls{dp} in comparison to others, and wish for the learner to sample more uniformly in cases of uncertainty (in terms of small differences between the \gls{dp} values).

Like \citet[p.~6]{eysenbach2019diversity}, we believe that \gls{vic}'s goal-selection policy collapses due to the choice of optimiser \citep[REINFORCE;][]{williams1992simple}. But our results, as well as those from the wider \gls{lp} literature, demonstrate their claim that learning $p(\rvg)$ ``results in learning fewer diverse skills'' (Appendix~E.2, p~18) to be overgeneralised.

\subsection{Evaluating skill diversity with dimensionality reduction}
\label{appendix:tsne}

To generate Figure~\ref{fig:tsne100epochs}, we first draw 100 trajectories from each skill, for each: (1) a set of random skills (no learning), (2) skills learned using \gls{diayn} (uniform goal selection), and (3) skills learned using \gls{diayn} with \gls{dp} motivating goal selection. Then, we compute the multivariate means over each trajectory, and dimension-reduce the feature space to two dimensions using \gls{tsne} with the following hyperparemeters: a perplexity of 30, a learning rate of 10, and running the optimisation for a maximum of 5000 iterations. In the plot, the distances between clusters may mean nothing \citep{wattenberg2016how}.

The same random seeds were used for the training of \gls{diayn} and \gls{diayn}+\gls{dp}. Similarly, the seeds for the random number generators of both the policy simulator and \gls{tsne} algorithm were fixed for the processes of simulating trajectories and dimensionality reduction across all three methods.

\end{document}